\documentclass[11pt,a4paper]{article}
\usepackage[hyperref]{emnlp2018}
\usepackage{times}
\usepackage{latexsym}
\usepackage{bbm}
\usepackage{multirow}
\usepackage{graphicx}
\usepackage{latexsym}
\usepackage{amsmath}
\usepackage{amsfonts}
\usepackage{multicol}
\usepackage{multirow}
\usepackage{color}
\usepackage{url}

\usepackage[colorinlistoftodos]{todonotes}
\usepackage{url}

\aclfinalcopy 

\newcommand*{\affaddr}[1]{#1} 
\newcommand*{\affmark}[1][*]{\textsuperscript{#1}}
\newcommand*{\email}[1]{\texttt{#1}}

\title{Improving Reinforcement Learning Based Image Captioning with Natural Language Prior}

\author{%
Tszhang Guo\affmark[1], Shiyu Chang\affmark[2], Mo Yu\affmark[2], Kun Bai\affmark[1]\\
\affaddr{\affmark[1]Mobile Internet Group, Tencent} \\
\affaddr{\affmark[2]MIT-IBM Watson AI Lab, IBM Research}\\
\email{\{kkzkguo,kunbai\}@tencent.com}\\
\email{shiyu.chang@ibm.com}, \email{yum@us.ibm.com}%
}
  
\date{}

\begin{document}
\maketitle
\begin{abstract}
Recently, Reinforcement Learning (RL) approaches have demonstrated advanced performance in image captioning by directly optimizing the metric used for testing.  However, this shaped reward introduces learning biases, which reduces the readability of generated text.  In addition, the large sample space makes training unstable and slow.  To alleviate these issues, we propose a simple coherent solution that constrains the action space using an $n$-gram language prior.  Quantitative and qualitative evaluations on benchmarks show that RL with the simple add-on module performs favorably against its counterpart in terms of both readability and speed of convergence. Human evaluation results show that our model is more human readable and graceful. The implementation will become publicly available upon the acceptance of the paper\footnote{https://github.com/tgGuo15/PriorImageCaption}.
\end{abstract}

\section{Introduction}
\label{sec:intro}
Image captioning \cite{Farhadi2010Every, Kulkarni2011Baby, 
Yao2016Boosting, Lu2016Knowing, 
Dai_2017_ICCV, Li2017Scene} aims at generating natural language descriptions of images.  Advanced by recent developments of deep learning, many captioning models rely on an encoder-decoder based paradigm \cite{Vinyals2015Show}, where the input image is encoded into hidden representations using a Convolutional Neural Network (CNN) followed by a Recurrent Neural Network (RNN) decoder to generate a word sequence as the caption.  Further, the decoder RNN can be equipped with spatial attention mechanisms \cite{Xu2015Show} to incorporate precise visual contexts, which often yields performance improvements empirically.

Although the encoder-decoder framework can be effectively trained with maximum likelihood estimation (MLE) \cite{Salakhutdinov2010Learning}, recent research \cite{Ranzato2015Sequence} have pointed out that the MLE based approaches suffer from the so-called exposure bias problem.  To address this problem, \cite{Ranzato2015Sequence}  proposed a Reinforcement Learning (RL) based training framework.  The method, developed on top of the REINFORCE algorithm \cite{williams1992simple}, directly optimizes the non-differentiable test metric (\emph{e.g.} BLEU \cite{papineni2002bleu}, CIDEr \cite{vedantam2015cider}, METEOR \cite{banerjee2005meteor} \emph{etc.}), and achieves promising improvements.   However, learning with RL is a notoriously difficult task due to the high-variance of gradient estimation.  Actor-critic \cite{sutton1998reinforcement} methods are often adopted, which involves training an additional value network to predict the expected reward.   On the other hand, \cite{Rennie2016Self} designed a self-critical method that utilizes the output of its own test-time inference algorithm as the baseline to normalize the rewards, which leads to further performance gains.

\begin{figure}[t!]
\centering
\includegraphics[width=0.48\textwidth]{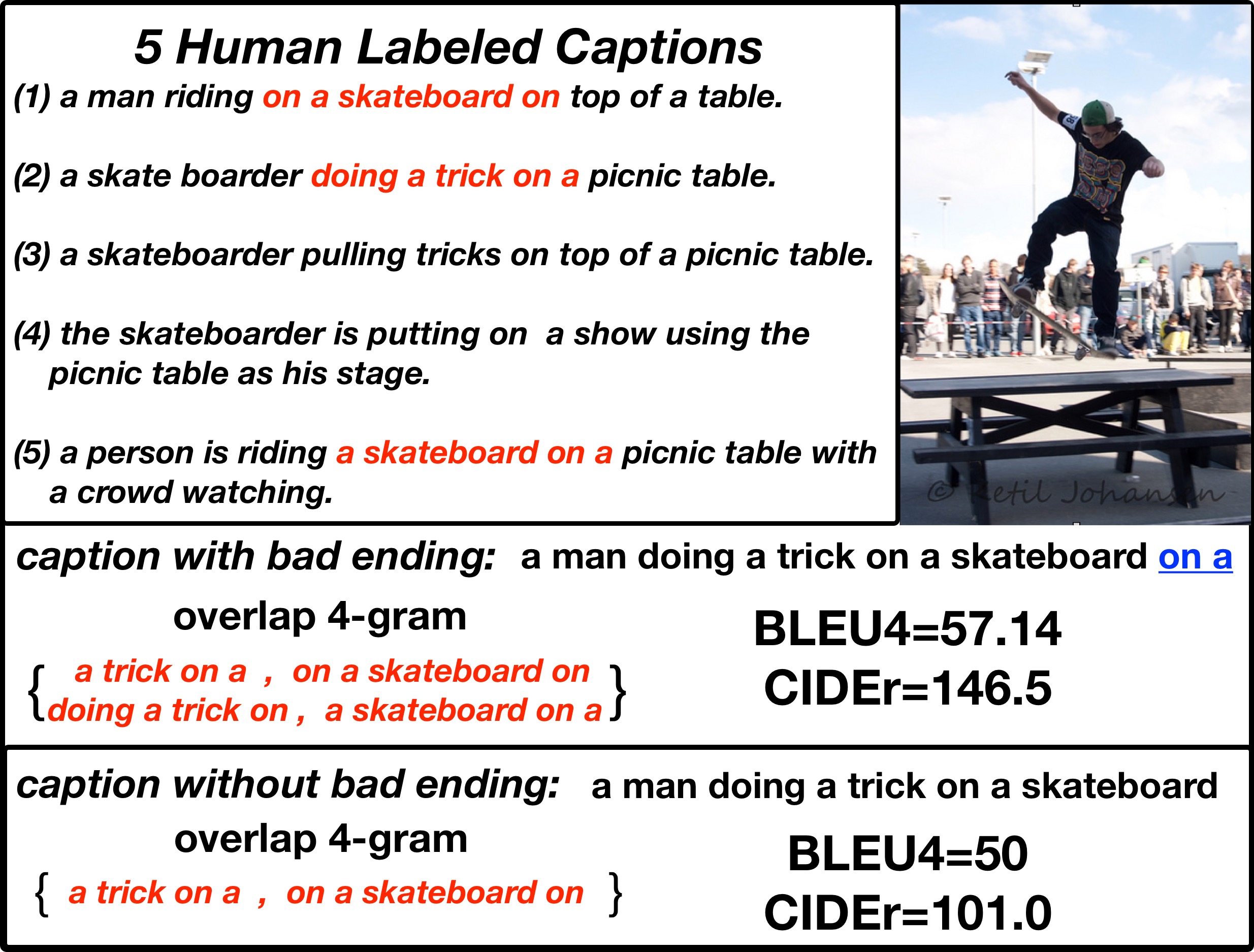}
\vspace{-0.3in}
\caption{A demonstration of the sequence with bad ending has higher BLEU and CIDEr scores compared to the one without. }
\label{fig:bad_end_metric}
\end{figure}

Beside to the high-variance problem, we notice that there are two other drawbacks of RL-based captioning methods that are often overlooked in the literature.  First, while these methods can directly optimize the non-differentiable rewards and achieve high test scores, the generated captions contain many repeated trivial patterns, especially at the end of the sequence.  Table \ref{tab:bad_cider} shows examples of bad-endings generated by a self-critical based RL algorithm (model details refer to Section \ref{sec:exp}).  Specifically, 46.44\% generated captions end with phrases as ``with a'', ``on a'', ``of a'', \emph{etc.} (for detailed statistics see Appendix \ref{sec:appA}), on the MSCOCO \cite{Chen2015Microsoft} validation set with the standard data splitting by \cite{Karpathy2015Deep}.  The reason is that the shaped reward function biases the learning.  In Figure \ref{fig:bad_end_metric}, we see these additive patterns at the end of captions, although make no sense to humans, yield to a higher reward.  Empirically, removing these endings results in a huge performance drop of around 6\%.  \cite{paulus2017deep} has also reported that in abstractive summarization, using RL only achieves high ROUGE \cite{lin2004rouge} score, yet the human-readability is very poor.  The second drawback is that RL-based text generation is sample-inefficient due to the large action space.  Specifically, the search space is of size $O(|\mathcal{V}|^T)$, where $\mathcal{V}$ is a set of words, $T$ is the sentence length, and $|\cdot|$ denotes the cardinality of a set.  This often makes training unstable and converge slowly.

In this work, to tackle these two issues, we propose a simple yet effective solution by introducing coherent language constraints on local action selections in RL.  Specifically, we first obtain word-level $n$-gram \cite{kneser1995improved} model from the training set and then use it as an effective prior.  During the action sampling step in RL, we reduce the search space of actions based on the constitution of the previous word contexts as well as our $n$-gram model.  To further promote samples with high rewards, we sample multiple sentences during the training and update the policy based on the best-rewarded one.   Such simple treatments prevent the appearance of bad endings and expedite the convergence while maintaining comparable performance to the pure RL counterpart.  In addition, the proposed framework is generic, which can be applied to many different kinds of neural structures and applications.

\begin{table}[t!]
\small
\setlength{\tabcolsep}{0.4mm}
\centering
\label{tab:bad_cider}
\begin{tabular}{lcc} \hline
Image ID & Generated sentence                        & CIDEr \\ \hline \hline
262262   & a tall building with a clock tower \textcolor{blue}{with a}  & 160.1 \\
262148   & a man doing a trick on a skateboard \textcolor{blue}{on a}  & 146.5 \\
52413 & a person holding a cell phone \textcolor{blue}{in a}  & 132.4 \\
393225   & a bow of soup with carrots \textcolor{blue}{and a}          & 118.5 \\ \hline
\end{tabular}
\vspace{-0.05in}
\caption{{Examples of bad sequences generated by a self-critical based RL baseline.  Blue color indicates the bad ending.  Sequences with bad endings have high CIDEr scores.}}
\end{table}


\section{Model Architecture}
\label{sec:model}
\paragraph{Encoder-Decoder Model:}
We adopt a similar structure as GNIC \cite{Vinyals2015Show}, which first encodes an image $I$ to a dense vector $h_I$ by CNN.   The vector $h_I$ is then fed as the input to an LSTM-based \cite{Hochreiter1997Long} language model decoder.  At each step $t$, the LSTM receives the previous output $w_{t-1}$ as the input; computes the hidden state $h_t$; and predicts the next word $w_t$ as below:
\begin{equation}
\label{eq:LSTM_decoder}
\begin{aligned}
& h_{t} = \text{LSTM}(h_{t-1}, w_{t-1}), ~~ l_{t} = W_{l}h_{t} \\
& w_{t} \sim \text{softmax}(l_{t}),
\end{aligned}
\end{equation}
where $w_0=h_I$ and $h_{0}$ and $c_{0}$ are initialized to zero. The generation ends if a special token *end* is predicted.

\paragraph{Attention Model:} Instead of utilizing a static representation of the image, attention mechanism dynamically reweights the spatial features from CNN to focus on the different region of the image at each word generation.  We specifically consider the standard architecture used in \cite{Xu2015Show}, where $\mathcal{A} = \{a_{1},a_{2},...,a_{L}\}$ is the spatial feature set and each $a_{i}\in R^{D}$ corresponds to features extracted at different image locations.  Then the hidden states of the LSTM is computed as
\begin{equation}
\begin{aligned}
& e_{ti} = f_{\text{att}}(a_{i}, h_{t-1})  , ~~ \beta_{ti} = \frac{\text{exp}(e_{ti})}{\sum_{k=1}^{L}\text{exp}(e_{tk})},\\
& z_{t} = \sum_{k=1}^{L} \beta_{tk} a_{k}, ~~ h_{t} = \text{LSTM}([h_{t-1},z_{t}], w_{t-1}),
\end{aligned}
\end{equation}
where $f_{\text{att}}$ is an attention model, which we use a single fully connected layer conditioned on the previous hidden state.  Once $h_t$ is obtained, the word generation is same as equation (\ref{eq:LSTM_decoder}).

\paragraph{Sequence Generation with RL: }We follow the training procedure of \cite{Rennie2016Self}.  The decoder LSTM can be viewed as a ``policy'' denoted by $p_{\theta}$, where $\theta$ is the set of parameters of the network.  At each time step $t$, the policy chooses an action by generating a word $w_t$ and obtains a new ``state'' (\emph{i.e.} hidden states of LSTM, attention weights, \emph{etc.}).  Once the end token is generated,  a ``reward'' $r$ is given based on the score (\emph{e.g.} CIDEr or BLEU) of the predicted sentence.  The goal is to maximize the expected reward as
\begin{equation}
L(\theta) = \mathbb{E}_{w^{s} \sim p_{\theta}} [r(w^{s})],
\end{equation}
where $w^{s} = \{w^{s}_{1}, w^{s}_{2},..., w^{s}_{T} \}$ are sampled words at every time step. The REINFORCE algorithm \cite{williams1992simple} provides unbiased gradient estimation of $\theta$ as
\begin{equation}
\nabla_{\theta}{L}(\theta) \approx r(w^{s}) \nabla_{\theta}{\log{p_{\theta}(w^{s})}},
\end{equation}
using a single sequence.

\paragraph{Variance Reduction with Self-Critical: }We reduce the variance of the gradient estimator by using the self-critical approach 
as
\begin{equation}
\nabla_{\theta}{L}(\theta) \approx (r(w^{s}) - r(\bar{w}))\nabla_{\theta}{\log{p_{\theta}(w^{s})}},
\end{equation}
where $\bar{w}_{t}$ is the baseline reward calculated by the current model under the inference algorithm used at test time defined as
\begin{equation}
\bar{w}_{t} = \mathop{\arg\max}_{w_{t}} p_{\theta}(w_{t}|h_{t}).
\end{equation}
Then, sequences have rewards higher than $\bar{w}$ will be increased in probability, while samples result in lower reward will be suppressed.

\section{Prior Language Constraint with $N$-Gram Model}

\paragraph{Method:} We collect all $n$-grams ($n$=3 or 4 in our experiments) from a corpus of captions. We use the training set from MSCOCO to avoid the usage of the additional resource.  Thus, a fair comparison to previous methods is guaranteed.  Then, we filter the $n$-grams with frequencies lower than five.  The set of remaining ones is denoted as $\mathcal{F}$.  During training, given the previous tokens predicted by the decoder, we constraint the sample space the current prediction by
\vspace{-0.05in}
\begin{equation}
\vspace{-0.05in}
w_{t} \sim \text{softmax}(p_{\theta}(w^{s}_{t}) \cdot \boldsymbol{\alpha_t}),
\end{equation}
where $\boldsymbol{\alpha_i}$ is an indicator vector whose length is the vocabulary size $\vert \mathcal{V} \vert$ and its elements are non-zero only if the corresponding word and the previous ($n-1$)-gram constitute a valid $n$-gram in $\mathcal{F}$ as
\begin{equation}
\small
\begin{aligned}
\boldsymbol{\alpha_t}[k]  =& \begin{cases}  1 &\mbox{if $\{w^{s}_{t-n+1}, \cdots, w^{s}_{t-1}, k\} \in \mathcal{F}$}\\ 0 &\mbox{otherwise} \end{cases}.
\end{aligned}
\end{equation}

\begin{table*}[t!]
\small
\centering
\begin{tabular}{ccccccc}
\hline
\multicolumn{1}{l}{}                                        & \bf Methods                      & \bf CIDEr         & \bf BLEU4       & \bf ROUGE-L       & \bf METEOR  &  \bf BadEnd-Rate    \\ \hline \hline
\multicolumn{1}{l}{\multirow{4}{*}{\rotatebox{90}{Published}}} & \cite{Karpathy2015Deep}          & 66.0          & 23.0        & - -         & 19.5   & 0.0     \\
\multicolumn{1}{l}{}                                        & \cite{Xu2015Show}                & - -           & 25.0        & - -         & 23.0    & 0.0    \\
\multicolumn{1}{l}{}                                        & MIXER \cite{Ranzato2015Sequence} & - -             & 29.1        & - -           & - -  & - -         \\
\multicolumn{1}{l}{}                                        & \cite{Zhou2017Deep}              & 93.7          & 30.4        & 52.5        & 25.1   & - -     \\ \hline \hline
\multirow{7}{*}{\rotatebox{90}{Implemented}}                & ED-XE                            & 89.8          & 28.0        & 51.7        & 24.2   & 0.0      \\
                                                            &
Att-XE                            & 95.1          & 29.2        & 52.8       & 24.8    & 0.0     \\
                                                            & ED-SC \cite{Rennie2016Self}      & 101.8 / \textcolor{blue}{\underline{96.1}}  & 31.2 / \textcolor{blue}{\underline{30.3}} & 53.1 / \textcolor{blue}{\underline{52.9}} & 24.6 / \textcolor{blue}{\underline{23.9}}   & 46.4\% / \textcolor{blue}{\underline{0.0}}       \\
                                                            & Att-SC \cite{Rennie2016Self}     & 105.7 /  \textcolor{blue}{\underline{100.8}} & 32.3 / \textcolor{purple}{\underline{30.8}} & 53.8 / \textcolor{blue}{\underline{53.1}} & 25.2 / \textcolor{blue}{\underline{24.1}}  &  43.7\% / \textcolor{blue}{\underline{0.0}} \\
                                                            & Ours-ED-4-gram                   & 96.7          & 29.1        & 51.4        & 23.9    & 0.0    \\
                                                            & Ours-Att-4-gram                  & \textcolor{red}{\bf{102.0}}    & 30.2        & \textcolor{red}{\bf{53.6}}   & \textcolor{red}{\bf{25.6}}  & 0.0 \\
                                                            & Ours-ED-tri-gram                 & 95.1          & 29.8        & 52.4        & 24.1    & 0.0    \\
                                                            & Ours-Att-tri-gram                & 100.4         & 28.7        & 51.8        & 25.0   & 0.0     \\ \hline\hline
                                                            
\end{tabular}
\vspace{-0.1in}
\caption{Quantitative evaluation of our method compared to baselines on MSCOCO.  \textcolor{blue}{\underline{Blue}} text indicates the performance after adjustments and \textbf{\textcolor{red}{red}} text indicates the best performance.}
\label{tab:performance}
\end{table*}

\begin{figure}
\centering
\includegraphics[width=0.5\textwidth]{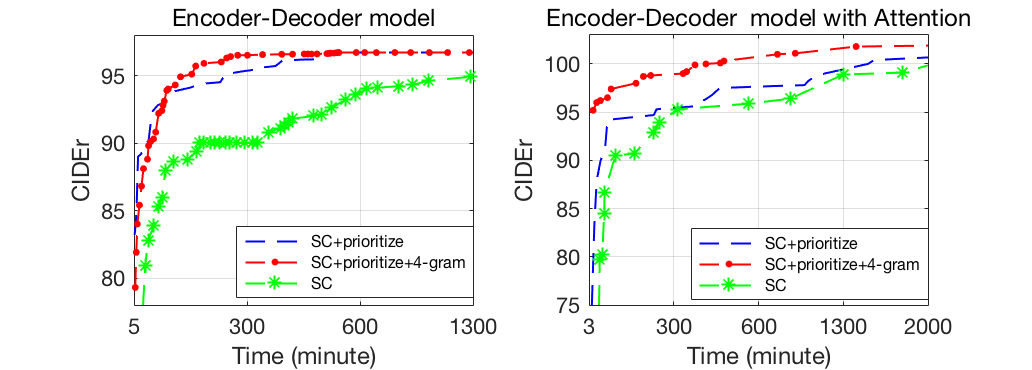}
\vspace{-0.2in}
\caption{Training time of models with (right) and without (left) spatial attention.} 
\label{fig:speed}
\end{figure}
\paragraph{Discussion: }The key motivation for applying the above constraint is two-fold: (1) this ensures generated captions always formed 
by valid $n$-grams, which provides us a direct way of eliminating the repeated common phrases and bad-endings like the ones in Table \ref{tab:bad_cider}; and (2) this shrinks the size of action space, which makes the training converges much faster.  For MSCOCO, action space is changed from more than 9,000 to 56 on average.


\section{Experiments}
\label{sec:exp}
\paragraph{Dataset: } We perform both quantitative and qualitative evaluations on MSCOCO dataset.  The dataset contains 123,287 images and each image has at least five human captions.  To seek fair comparison to others, we use the publicly available splits, which contains 82,783 training, 5,000 validation and 5,000 testing images.
\paragraph{Implementation Details: } Our implementations are based on the publicly
project.\footnote{https://github.com/ruotianluo/self-critical.pytorch} 
We use an ImageNet pre-trained 101-layered ResNet\footnote{https://github.com/KaimingHe/deep-residual-networks} \cite{He2016Deep} to extract visual features. We consider two types (see Section \ref{sec:model}) of architectural training with RL: (1) the plain encoder-decoder, and (2) the encoder-decoder with attention.  For the former one, we represent each image by a 2,048-dimension vector by extracting the features from the last convolutional layer with average pooling.  For the attention model,  we apply spatial adaptive max pooling and the output feature map has the size of $14 \times 14 \times 2,048$.  At each time step, the attention model produces weights over 196 spatial locations.  The size of word embeddings and the hidden dimension of the LSTM are set to 512 for all experiments. More details are in Appendix \ref{sec:appB}.

\paragraph{Compared Methods: } We report our results in four different settings, which include the combinations of with/without attention and using tri-/four-gram.  We directly compare with our counterparts that have the same structures but no $n$-gram modules.  Specifically, they are encoder-decoder based self-critical (ED-SC), and the one with attention (Att-SC).  In addition, since our experimental setup is almost identical to many existing works, we also include their reported results, which include \cite{Karpathy2015Deep, Xu2015Show, Ranzato2015Sequence, Zhou2017Deep}.  At last, we also include the performance of our warm-start models - the models trained by MLE \cite{Vinyals2015Show} using cross entropy (ED-XE and Att-XE) - as a reference.
\paragraph{Evaluation Metric and Performance Adjustment: }We report performance on FIVE metrics: BLEU4, METEOR, ROUGE-L,CIDEr and Bad Ending Rate.  For the self-critical baselines, we report two sets of performances: 1) the captions directly generated by the model; and 2) the sequences of removing bad endings of the generated captions, based on the distribution in Appendix \ref{sec:appA}.  

\paragraph{Results:} Table \ref{tab:performance} summarizes the performances of our models compared with other baselines.  We see that without performance adjustments, the self-critical RL with attention performs the best.  However, since it contains many bad endings, our method achieves supreme results after these repeated patterns are removed.   We also provide some qualitative comparison between our attention model and self-critical in Appendix \ref{sec:appC}.
\paragraph{Efficient Training: } We show that constraining the action space leads to a more efficient RL training in Figure \ref{fig:speed}. CIDEr score is calculated after removing bad endings.  We plot three curves using architectures with/without attentions.  The Green curve is the self-critical, the blue one is with prioritized sampling, and the red one is our final model with 4-gram constraint.  We observe that we can speed up almost twice than its counterpart.

\paragraph{Online Evaluation:} We also evaluate our attention model on COCO online server\footnote{https://competitions.codalab.org/competitions/3221} and results are reported in Table \ref{Onlie}. Att-SC gets a higher score than ours in the online test, however, with a lot of bad endings where the bad ending ratio is 72.7$\%$.
\begin{table*}[t]
\small
\centering
\begin{tabular}{cccccc}
\hline \bf Methods                      & \bf CIDEr         & \bf BLEU4       & \bf METEOR        & \bf ROUGE-L  &  \bf BadEnd-Rate \\
\hline \hline
Att-SC &\textcolor{red}{\bf{109.3}}  & \textcolor{red}{\bf{61.9}} & 32.9& 67.7  &   72.7\% \\ 
Att-4-gram &104.7 & 59.8 & 33.0 & 66.2 & 0 \\ 
Att-LSTM-LM &104.3 & 61.0 & \textcolor{red}{\bf{33.9}}  & \textcolor{red}{\bf{68.5}}   & 0\\ \hline
\hline
\end{tabular}
\vspace{-0.1in}
\caption{Quantitative results on online server (C40 test).\textbf{\textcolor{red}{Red}} text indicates the best performance.}
\label{Onlie}
\end{table*}

\begin{table*}[t!]
\small
\centering
\begin{tabular}{cccccc}
\hline
 \bf Methods                      & \bf CIDEr         & \bf BLEU4       & \bf ROUGE-L       & \bf METEOR  &  \bf BadEnd-Rate    \\ \hline \hline

Att-SC \cite{Rennie2016Self}     & 105.7 /  \textcolor{blue}{\underline{100.8}} & 32.3 / \textcolor{blue}{\underline{30.8}} & 53.8 / \textcolor{blue}{\underline{53.1}} & 25.2 / \textcolor{blue}{\underline{24.1}}  &  43.7\% / \textcolor{blue}{\underline{0.0}} \\
 Ours-ED-4-gram                   & 96.7          & 29.1        & 51.4        & 23.9    & 0.0  \\
 Ours-Att-4-gram                  & 102.0    & 30.2        & 53.6   & \textcolor{red}{\bf{25.6}}  & 0.0 \\ \hline
   ED-LSTM-LM                 & 99.4          & 30.9        & 52.7        & 24.6    & 0.0    \\
 Att-LSTM-LM                &  \textcolor{red}{\bf{105.9}}        & \textcolor{red}{\bf{32.8}}       &  \textcolor{red}{\bf{54.1}}       & 25.4   & 0.0     \\\hline \hline
\end{tabular}
\vspace{-0.1in}
\caption{Quantitative evaluation with our extension methods on MSCOCO.  \textcolor{blue}{\underline{Blue}} text indicates the performance after adjustments and \textbf{\textcolor{red}{red}} text indicates the best performance.}
\label{tab:Extension}
\end{table*}

\paragraph{Human Evaluation:} We also implement human evaluation on the results generated by our Att-4-gram compared with Att-SC. We randomly select 200 images from the test set. Each time, one image with two captions generated by two different models are shown to the volunteer and three choices are provided: (1) the first one is better; (2) both are the same level; (3) the second one is better. See more details in Appendix \ref{sec:appD}. In Table \ref{turing}, our model wins 400 times and performs more closely to human than Att-SC.
\begin{table} [t]
\vspace{-0.05in}
\small
\centering
\begin{tabular}{cccc}
\hline Methods & 4-gram win & Same level & 4-gram lose\\
\hline
4-gram \textbf{VS}  SC &  400 & 349 & 251  \\ \hline
\end{tabular}
\vspace{-0.1in}
\caption{Human evaluation results for attention models}
\label{turing}
\vspace{-0.05in}
\end{table}

\paragraph{Evaluating Captions Diversity:} To further evaluate the quality of the caption model, we follow \cite{Shetty2017Speaking} to measure the diversity of the generated captions.  We compute the novelty score of our 4-gram model, which is defined as whether a particular caption has been observed in the training set.   When two models have the same level predictive performances (\emph{e.g.} CIDEr), a higher novelty score usually indicates more diverse generations.   We conduct the experiment five times and report the averaged novelty score of our 4-gram model and the Att-SC, which are 77.83\% and 59.28\% respectively.   As the reference, the METEOR and novelty scores reported in \cite{Shetty2017Speaking} are 23.6, and 79.84\%, respectively.



\section{Neural Language Models Extension}
Inspired by the paper reviews, we extend our model by adopting another language prior to evaluating the effectiveness of constraining action space during REINFORCE training. We train our neural language model based on the MSCOCO caption corpus with an LSTM unit.

\paragraph{LSTM Language Model:} Given a word series $\{w_{0},w_{1},...,w_{T}\}$, the target of a neural language model is to maximize the log-likelihood as:
\begin{equation}
\label{eq:LSTM_lm}
\begin{aligned}
 \mathop{\max}_{\theta} \log p_{\theta}(w_{0},w_{1},...,w_{T}). 
\end{aligned}
\end{equation}
We model $p_{\theta}(w_{0},w_{1},...,w_{T})$ by an LSTM unit:
\begin{equation}
\label{eq:LSTM_lm}
\begin{aligned}
\log p_{\theta}(&w_{0}, ...,w_{T})= \sum_{t=1}^{T} \log p_{\theta_{LM}}(w_{t}|h_{t-1}) \\
&h_{t} = \text{LSTM}_{LM}(h_{t-1},w_{t-1}), 
\end{aligned}
\end{equation}
where $w_{0}$ is set to a *start* token for all sentences. $h_{0}$ and $c_{0}$ are initialized to zero.  After obtaining the optimized $\theta^{*}_{LM}$, we can use it to constrain the action space similar to the N-gram language model.  Specifically, given previous $t-1$ sampled words from current caption model, we compute $p_{\theta^{*}_{LM}}(w_{t} | w_{0},w_{1},...,w_{t-1})$, which is the probability of the next word over the entire vocabulary.  We then apply a simple thresholding rule to form a subset of valid words for the captioning model.  
\begin{equation}
\begin{aligned}
& \boldsymbol{\alpha_t}[k]  = \begin{cases}  1 &\mbox{if $\{k\} \in \mathcal{F}$}\\ 0 &\mbox{otherwise} \end{cases}, \text{where} \\
 \mathcal{F} =& \{w_{t}|p_{\theta^{*}_{LM}}(w_{t}|w_{0},w_{1},...,w_{t-1}) \geq \eta \}.
\end{aligned}
\end{equation}
$\eta$ is a hyperparameter.  

\paragraph{Additional Experiments} 
The word embedding size and hidden dimension of $\theta_{LM}$ are set to 256 for this experiment.  We use Adam optimizer for training language model and the learning rate is set to 0.001. The batch size of language model training and REINFORCE training are both set to 20 in the experiments.  $\eta$ is set to 0.00005 for the first word and increases by a factor of two for every timestep.  We report our results in two settings, which include the combination of with/without attention for the caption model (termed ED-LSTM-LM and Att-LSTM-LM). We use the same warm-start models as in the N-gram experiments. The performances are summarized in Table \ref{tab:Extension} and Table \ref{Onlie}.  We see that the neural language model provides further performance gains compared to the N-gram model without introducing any bad-endings.   
This is because that the LSTM language model covers a larger context than N-gram, which helps to generate more accurate captions.




\section{Conclusion}
In this paper, we present a simple but efficient approach to RL-based image caption by considering $n$-gram language prior to constrain the action space.  Our method converges faster and achieves better results than self-critical setting after removing bad endings in the generated captions.  In addition, captions generated by our models are more human readable and graceful.  We further extend our ideas using neural language model.   The results demonstrate that the captioning models are more beneficial from the neural language model than the N-gram model.


\newpage
\bibliography{emnlp2018}

\begin{thebibliography}{27}
\expandafter\ifx\csname natexlab\endcsname\relax\def\natexlab#1{#1}\fi

\bibitem[{Banerjee and Lavie(2005)}]{banerjee2005meteor}
Satanjeev Banerjee and Alon Lavie. 2005.
\newblock Meteor: An automatic metric for mt evaluation with improved
  correlation with human judgments.
\newblock In \emph{ACL-workshop}, pages 228--231.

\bibitem[{Chen et~al.(2015)Chen, Fang, Lin, Vedantam, Gupta, Dollar, and
  Zitnick}]{Chen2015Microsoft}
Xinlei Chen, Hao Fang, Tsung~Yi Lin, Ramakrishna Vedantam, Saurabh Gupta, Piotr
  Dollar, and C.~Lawrence Zitnick. 2015.
\newblock Microsoft coco captions: Data collection and evaluation server.
\newblock \emph{Computer Science}.

\bibitem[{Dai et~al.(2017)Dai, Fidler, Urtasun, and Lin}]{Dai_2017_ICCV}
Bo~Dai, Sanja Fidler, Raquel Urtasun, and Dahua Lin. 2017.
\newblock Towards diverse and natural image descriptions via a conditional gan.
\newblock In \emph{ICCV}, pages 2989--2998.

\bibitem[{Farhadi et~al.(2010)Farhadi, Hejrati, Sadeghi, Young, Rashtchian,
  Hockenmaier, and Forsyth}]{Farhadi2010Every}
Ali Farhadi, Mohsen Hejrati, Mohammad~Amin Sadeghi, Peter Young, Cyrus
  Rashtchian, Julia Hockenmaier, and David Forsyth. 2010.
\newblock Every picture tells a story: generating sentences from images.
\newblock \emph{Lecture Notes in Computer Science}, 21(10):15--29.

\bibitem[{He et~al.(2016)He, Zhang, Ren, and Sun}]{He2016Deep}
Kaiming He, Xiangyu Zhang, Shaoqing Ren, and Jian Sun. 2016.
\newblock Deep residual learning for image recognition.
\newblock In \emph{CVPR}, pages 770--778.

\bibitem[{Hochreiter and Schmidhuber(1997)}]{Hochreiter1997Long}
Sepp Hochreiter and Jürgen Schmidhuber. 1997.
\newblock Long short-term memory.
\newblock \emph{Neural Computation}, 9(8):1735--1780.

\bibitem[{Karpathy and Li(2015)}]{Karpathy2015Deep}
Andrej Karpathy and Fei~Fei Li. 2015.
\newblock Deep visual-semantic alignments for generating image descriptions.
\newblock In \emph{CVPR}, pages 3128--3137.

\bibitem[{Kingma and Ba(2015)}]{Kingma2014Adam}
Diederik Kingma and Jimmy Ba. 2015.
\newblock Adam: A method for stochastic optimization.
\newblock In \emph{ICLR}.

\bibitem[{Kneser and Ney(1995)}]{kneser1995improved}
Reinhard Kneser and Hermann Ney. 1995.
\newblock Improved backing-off for m-gram language modeling.
\newblock In \emph{Acoustics, Speech, and Signal Processing, 1995. ICASSP-95.,
  1995 International Conference on}, volume~1, pages 181--184. IEEE.

\bibitem[{Kulkarni et~al.(2011)Kulkarni, Premraj, Dhar, Li, Choi, Berg, and
  Berg}]{Kulkarni2011Baby}
G.~Kulkarni, V.~Premraj, S.~Dhar, Siming Li, Yejin Choi, A.~C. Berg, and T.~L.
  Berg. 2011.
\newblock Baby talk: Understanding and generating simple image descriptions.
\newblock In \emph{CVPR}, pages 1601--1608.

\bibitem[{Kusner and Hernándezlobato(2016)}]{Kusner2016GANS}
Matt~J. Kusner and José~Miguel Hernándezlobato. 2016.
\newblock Gans for sequences of discrete elements with the gumbel-softmax
  distribution.

\bibitem[{Li et~al.(2017)Li, Ouyang, Zhou, Wang, and Wang}]{Li2017Scene}
Yikang Li, Wanli Ouyang, Bolei Zhou, Kun Wang, and Xiaogang Wang. 2017.
\newblock Scene graph generation from objects, phrases and region captions.
\newblock In \emph{ICCV}, pages 1270--1279.

\bibitem[{Lin(2004)}]{lin2004rouge}
Chin-Yew Lin. 2004.
\newblock Rouge: A package for automatic evaluation of summaries.
\newblock In \emph{ACL-workshop}, page~10.

\bibitem[{Lu et~al.(2016)Lu, Xiong, Parikh, and Socher}]{Lu2016Knowing}
Jiasen Lu, Caiming Xiong, Devi Parikh, and Richard Socher. 2016.
\newblock Knowing when to look: Adaptive attention via a visual sentinel for
  image captioning.
\newblock pages 3242--3250.

\bibitem[{Papineni et~al.(2002)Papineni, Roukos, Ward, and
  Zhu}]{papineni2002bleu}
Kishore Papineni, Salim Roukos, Todd Ward, and Wei-Jing Zhu. 2002.
\newblock Bleu: a method for automatic evaluation of machine translation.
\newblock In \emph{ACL}, pages 311--318.

\bibitem[{Paulus et~al.(2017)Paulus, Xiong, and Socher}]{paulus2017deep}
Romain Paulus, Caiming Xiong, and Richard Socher. 2017.
\newblock A deep reinforced model for abstractive summarization.
\newblock \emph{arXiv preprint arXiv:1705.04304}.

\bibitem[{Ranzato et~al.(2015)Ranzato, Chopra, Auli, and
  Zaremba}]{Ranzato2015Sequence}
Marc'Aurelio Ranzato, Sumit Chopra, Michael Auli, and Wojciech Zaremba. 2015.
\newblock Sequence level training with recurrent neural networks.
\newblock \emph{Computer Science}.

\bibitem[{Ren et~al.(2017)Ren, Wang, Zhang, Lv, and Li}]{Zhou2017Deep}
Zhou Ren, Xiaoyu Wang, Ning Zhang, Xutao Lv, and LiJia Li. 2017.
\newblock Deep reinforcement learning-based image captioning with embedding
  reward.
\newblock In \emph{CVPR}, pages 1151--1159.

\bibitem[{Rennie et~al.(2017)Rennie, Marcheret, Mroueh, Ross, and
  Goel}]{Rennie2016Self}
Steven~J Rennie, Etienne Marcheret, Youssef Mroueh, Jarret Ross, and Vaibhava
  Goel. 2017.
\newblock Self-critical sequence training for image captioning.
\newblock In \emph{CVPR}, pages 1179--1195.

\bibitem[{Salakhutdinov(2010)}]{Salakhutdinov2010Learning}
Ruslan Salakhutdinov. 2010.
\newblock Learning deep generative models.
\newblock 2(1):361--385.

\bibitem[{Shetty et~al.(2017)Shetty, Rohrbach, Hendricks, Fritz, and
  Schiele}]{Shetty2017Speaking}
Rakshith Shetty, Marcus Rohrbach, Lisa~Anne Hendricks, Mario Fritz, and Bernt
  Schiele. 2017.
\newblock Speaking the same language: Matching machine to human captions by
  adversarial training.
\newblock In \emph{ICCV}, pages 4155--4164.

\bibitem[{Sutton and Barto(1998)}]{sutton1998reinforcement}
Richard~S Sutton and Andrew~G Barto. 1998.
\newblock \emph{Reinforcement learning: An introduction}, volume~1.
\newblock MIT press Cambridge.

\bibitem[{Vedantam et~al.(2015)Vedantam, Lawrence~Zitnick, and
  Parikh}]{vedantam2015cider}
Ramakrishna Vedantam, C~Lawrence~Zitnick, and Devi Parikh. 2015.
\newblock Cider: Consensus-based image description evaluation.
\newblock In \emph{CVPR}, pages 4566--4575.

\bibitem[{Vinyals et~al.(2015)Vinyals, Toshev, Bengio, and
  Erhan}]{Vinyals2015Show}
Oriol Vinyals, Alexander Toshev, Samy Bengio, and Dumitru Erhan. 2015.
\newblock Show and tell: A neural image caption generator.
\newblock In \emph{CVPR}, pages 3156--3164.

\bibitem[{Williams(1992)}]{williams1992simple}
Ronald~J Williams. 1992.
\newblock Simple statistical gradient-following algorithms for connectionist
  reinforcement learning.
\newblock In \emph{Reinforcement Learning}, pages 5--32. Springer.

\bibitem[{Xu et~al.(2015)Xu, Ba, Kiros, Courville, Salakhutdinov, Zemel, and
  Bengio}]{Xu2015Show}
Kelvin Xu, Jimmy Ba, Ryan Kiros, Aaron Courville, Ruslan Salakhutdinov, Richard
  Zemel, and Yoshua Bengio. 2015.
\newblock Show, attend and tell: Neural image caption generation with visual
  attention.
\newblock In \emph{ICML}, pages 2048--2057.

\bibitem[{Yao et~al.(2017)Yao, Pan, Li, Qiu, and Mei}]{Yao2016Boosting}
Ting Yao, Yingwei Pan, Yehao Li, Zhaofan Qiu, and Tao Mei. 2017.
\newblock Boosting image captioning with attributes.
\newblock In \emph{ICCV}, pages 4904--4912.

\end{thebibliography}
\bibliographystyle{acl_natbib_nourl}

\appendix

\clearpage
\section{Statistics of Bad Endings}
\label{sec:appA}

\begin{figure}[h]
\centering
\includegraphics[width=0.48\textwidth]{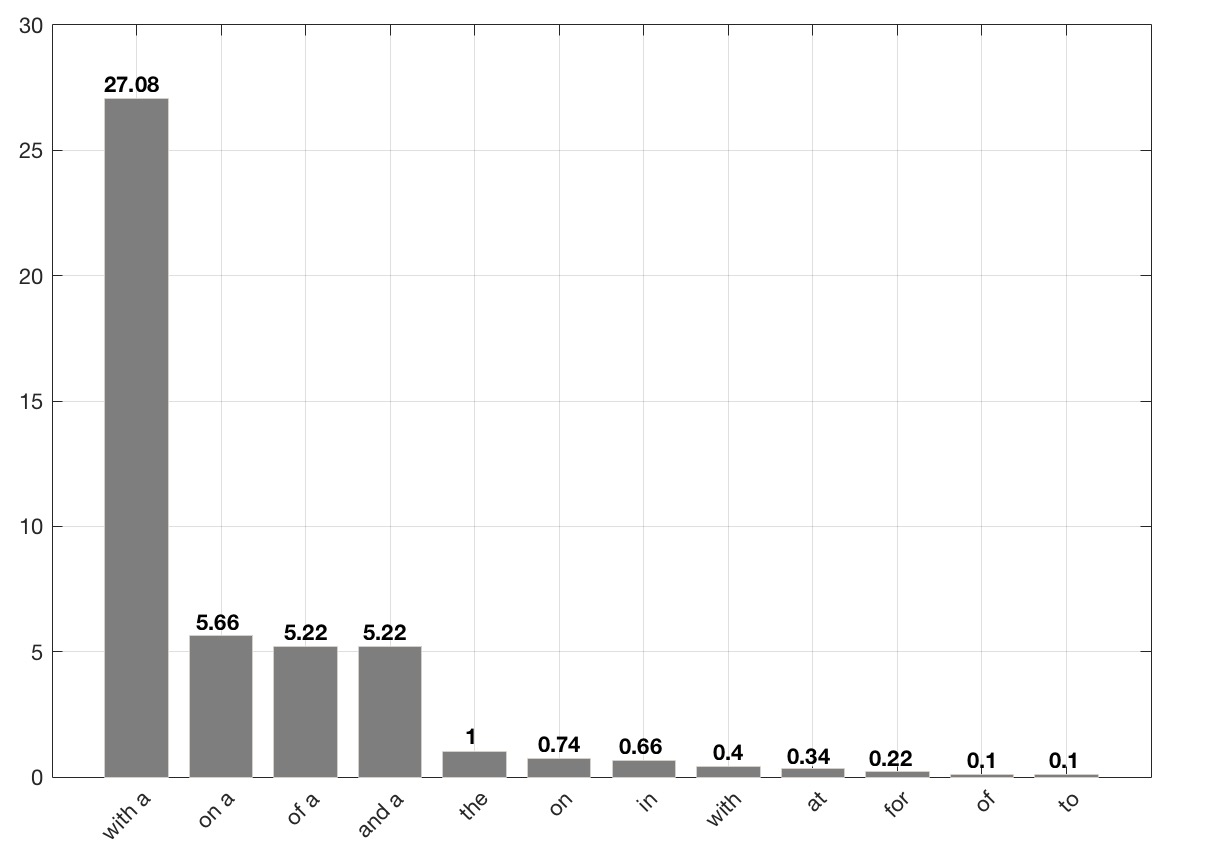}
\caption{Statistics of bad endings over 5,000 validation instances generated by self-critical method optimized on the CIDEr metric.}
\label{fig:bad_end_bar}
\end{figure}

\section{More training details}
\label{sec:appB}

We drop any words that have appeared less than five times.  The vocabulary size is 9,488.  We do not rescale or crop the images when extracting CNN features for the attention model. At the beginning of RL training, the learning rate is $5 \times 10^{-5}$ and we anneal it by a factor of 0.2 when the CIDEr score on validation set has no improvement for over 10 epochs. The CNN weights are fixed during our RL training process. We use a Gumbel sampler and perform our action sampling on GPU devices which is much faster than CPU device. The batch size is set to 50 in all our experiments. In our observation, we discover that a very powerful warm-start model is necessary to maintain the stability and convergence speed of self-critical without our prioritized sampling and n-gram constraints while our methods need not.

We warm-start all models by training them under the cross-entropy objective.  We use ADAM \cite{Kingma2014Adam} optimizer with an initial learning rate of $5 \times 10^{-4}$. We select the model with best CIDEr scores on the development set to initialize RL training and use a Gumbel sampler \cite{Kusner2016GANS} to improve action sampling efficiency. In order to promote captions with higher reward, we sample multiple sequences (we set 10 in our experiments) during the training and update the parameters based on the sample with the highest rewards.  We find that this technique empirically helps convergence.

\newpage

\section{Qualitative Comparison}
\label{sec:appC}

\begin{figure}[h!]
\centering
\includegraphics[width=0.5\textwidth]{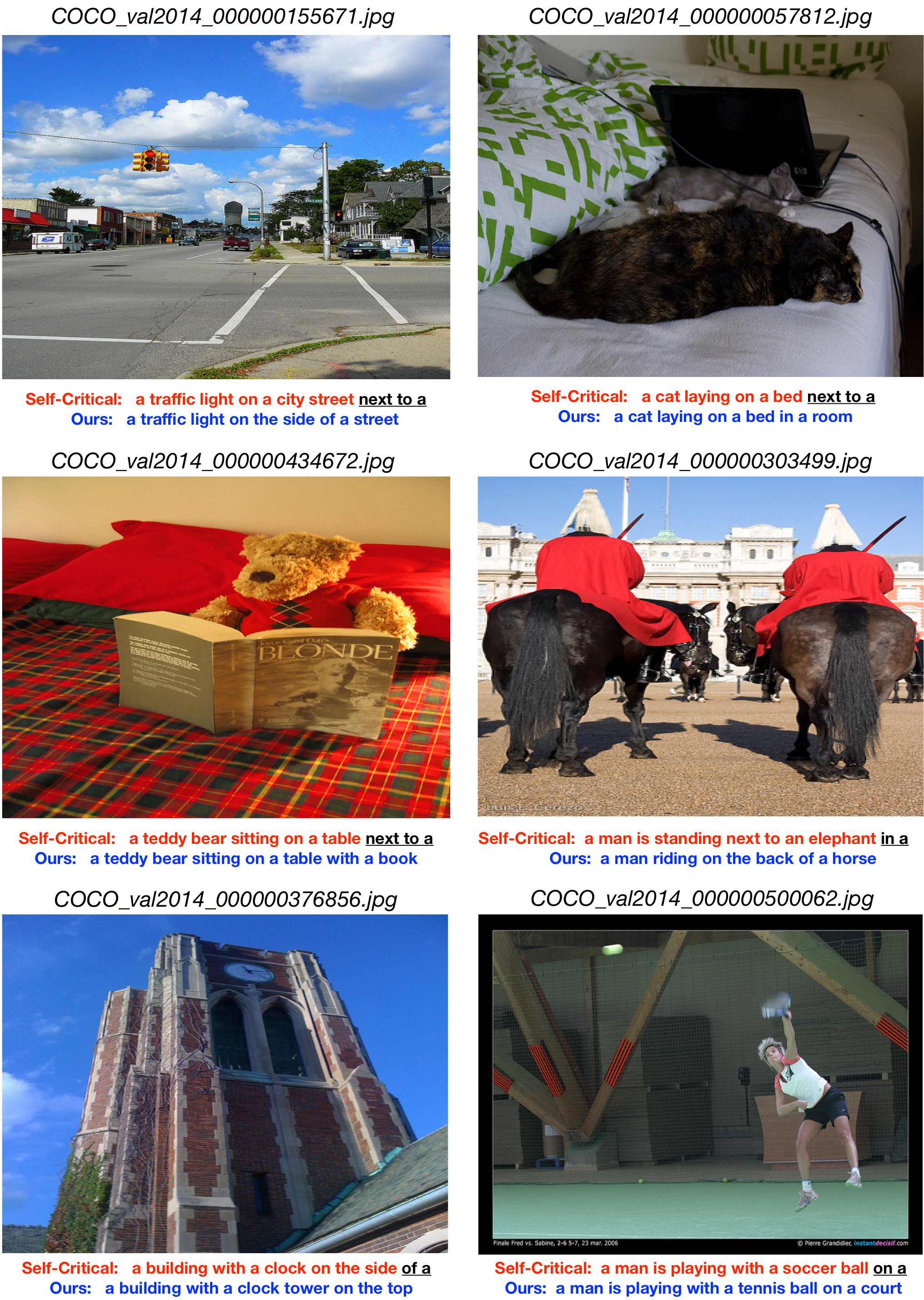}
\caption{Examples generated by our model with attention compared to the self-critical counterpart. After adding N-gram constraints, our results are more accurate and human-readable.}
\end{figure}

\section{Human Evaluation Details}
\label{sec:appD}
\begin{figure}[h!]
\centering
\includegraphics[width=0.3\textwidth]{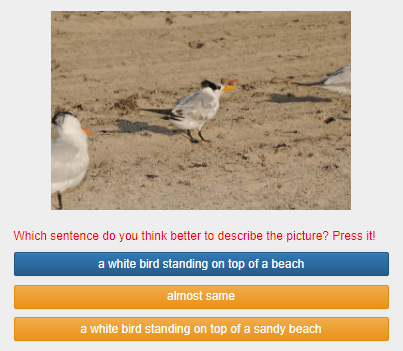}
\caption{Web page for human evaluation}
\label{webpage}
\end{figure}

We recruit 10 volunteers who are under the correct guidance for finishing the human evaluation process.We implement our human evaluation experiment on a web page (see Figure \ref{webpage}). 

Every time the volunteer must make a choice among these three choices. And "almost same" means that two captions are both good or both bad for the given images.  Each image must be evaluated by at least 5 volunteers.

The order of caption generated from given model is random so the volunteers have no idea where these captions come from which model. Some times the first one comes from Ours-Att-4-gram, and some times from Att-SC. 
\end{document}